\documentclass[runningheads]{llncs}
\usepackage[T1]{fontenc}
\usepackage{graphicx}
\usepackage{hyperref}
\usepackage{amsmath}
\usepackage{color}
\usepackage{subcaption}

\newcommand\toprule{\hline\noalign{\vspace{1mm}}}
\newcommand\midrule{\noalign{\vspace{1mm}}\hline\noalign{\vspace{1mm}}}
\newcommand\bottomrule{\noalign{\vspace{1mm}}\hline}

\begin{document}

\title{Interpretable Prediction of Lymph Node Metastasis in Rectal Cancer MRI Using Variational Autoencoders} 
\titlerunning{Interpretable LNM prediction}

\author{Benjamin Keel\inst{1}\orcidID{0000-0003-2763-7063} 
\and Aaron Quyn\inst{1,2}\orcidID{0000-0002-4333-5658} \and David Jayne\inst{1,2}\orcidID{0000-0002-8725-3283} \and Maryam Mohsin\inst{2} \and Samuel D. Relton\inst{1}\orcidID{0000-0003-0634-4587}
}

\authorrunning{B. Keel et al.}
\institute{University of Leeds, Leeds, UK\\
\email{mm17b2k@leeds.ac.uk}\\
\and Leeds Teaching Hospitals NHS Trust, Leeds, UK}

\maketitle              

\begin{abstract}
Effective treatment for rectal cancer relies on accurate lymph node metastasis (LNM) staging. However, radiological criteria based on lymph node (LN) size, shape and texture morphology have limited diagnostic accuracy. In this work, we investigate applying a Variational Autoencoder (VAE) as a feature encoder model to replace the large pre-trained Convolutional Neural Network (CNN) used in existing approaches. The motivation for using a VAE is that the generative model aims to reconstruct the images, so it directly encodes visual features and meaningful patterns across the data. This leads to a disentangled and structured latent space which can be more interpretable than a CNN. Models are deployed on an in-house MRI dataset with 168 patients who did not undergo neo-adjuvant treatment. The post-operative pathological N stage was used as the ground truth to evaluate model predictions. Our proposed model `VAE-MLP' achieved state-of-the-art performance on the MRI dataset, with cross-validated metrics of AUC $0.86 \pm 0.05$, Sensitivity $0.79 \pm 0.06$, and Specificity $0.85 \pm 0.05$. Code is available at: \url{https://github.com/benkeel/Lymph_Node_Classification_MIUA}.

\end{abstract}

\section{Introduction}
Rectal Cancer (RC) is the fifth most prevalent cancer in the UK and remains one of the leading causes of cancer related mortality, with a 5-year survival rate of 59.6\% \cite{nhs_cancer_stats}. Lymph node metastasis (LNM) is one of the most critical prognostic factors, as malignant lymph nodes (LNs) increase the risk of cancer recurrence and the development of distant metastasis \cite{valentini2011nomograms}. Consequently, LN staging informs the appropriate use of neo-adjuvant chemotherapy and radiotherapy. 
In clinical practice, radiologists use criteria based on size, shape and texture morphology to detect RC LNM on MRI \cite{doi:10.1148/rg.2019180114}. However, the established rules-based criteria have limited diagnostic sensitivity and specificity of 73\% (95\% CI: 68-77\%) and 74\% (95\% CI: 68-80\%) respectively \cite{Zhuang2021}. Inaccurate LNM staging can result in under or over treatment which can cause patient toxicity and affect cancer outcomes \cite{Borgheresi2022,doi:10.1148/rg.2019180114}. Compounding this, shortages of specialist radiologists have put significant pressure on hospitals and multidisciplinary teams (MDT) for planning cancer treatment, underlining the need for decision support tools.

Recent studies have demonstrated the capability of deep learning methods for accurately staging LNM on pre-operative radiologic imaging \cite{keel2024state-of-the-art}. However, a significant barrier to clinical adoption remains the lack of interpretability and robust validation of the models. This study investigates the application of Variational Autoencoders (VAEs) to predict LNM in RC. Proposed models were trained using 2D patches of LNs extracted from pre-operative MRI. Feature representations of the patches produced by the VAE were used to train Multi-layer Perceptron (MLP) classification models in a Multiple Instance Learning (MIL) framework. The approach used an MLP model to refine the feature representations into individual LN predictions, which were combined with clinical data in a second MLP for a patient level diagnosis. The final binary predictions were evaluated against the post-operative pathology confirmed N stage.
The novel contributions of this work are as follows:
\begin{enumerate}
    \item First study to exceed the diagnostic accuracy of radiologists using non-specialist segmentations of the lymph nodes on RC MRI.
    \item First application of generative AI in colorectal cancer LNM prediction, with the novel use of VAEs to improve the interpretability and demonstrate that the model latent space captures clinically meaningful information.
\end{enumerate}

\section{Related Works}
\label{sec:related_works}
In our scoping review of deep learning methods applied to pre-operative colorectal cancer LNM prediction (2018-2024), we reported a mean AUC of 0.856 (95\% CI: 0.796-0.916), outperforming radiologists and methods using radiomics features \cite{keel2024state-of-the-art}.  
However, existing studies have methodological limitations, including selection bias and restrictive inclusion criteria in study cohorts. Most approaches do not incorporate multi-modal clinical data and often include patients across multiple T-stages without using it as a variable in the model. Some results may also be inflated due to common weaknesses in the validation such as small test sets, reporting limited metrics, and only one previous study using cross-validation \cite{Xie2023}. There is also a general lack of explainability techniques beyond Grad-CAM \cite{Selvaraju_2019} to indicate regions of the image that have high importance to the diagnosis. 

A recent study by Xia et al. \cite{Xia2024_LNM} employed automatic LN detection and a 2D ResNet CNN classification architecture with pre-operative LN MRI patches alongside size characteristics and the apparent diffusion coefficient. Models were trained and evaluated on 1014 patients across three centres, including two external validation cohorts. They reported an AUC of 0.81, outperforming junior radiologists (0.69 (95\% CI: 0.64-0.73)) and senior radiologists (0.79 (95\% CI: 0.75-0.83)). Collaboration with the model significantly improved the performance of the radiologist groups to 0.80 (95\% CI: 0.76-0.84) and 0.88 (95\% CI: 0.85-0.91) respectively, demonstrating the value of RC LNM decision support tools to assist radiologists in clinical practice. This study was the most methodologically similar to ours and served as a primary reference.

The study methodology applying VAEs in medical image representation learning has been shown to be effective in a variety of diagnostic tasks \cite{10.1093/bioadv/vbac100,Redekop_2024_CVPR,10577275}. Several studies have shown that VAEs exceed the performance of CNNs in cancer imaging tasks, including state-of-the-art methods for pre-operative detection of breast cancer \cite{Sreelekshmi2024} and lung cancer \cite{VAE_lung_lesion} on public datasets. Another study compared VAE and CNN models for the pathological diagnosis of rectal cancer and found that the VAE was the best performing on the in-house dataset \cite{Kaur2023}.

Adjacent topics highlight that VAEs generate more meaningful and structured latent spaces and can provide more robustness in classification.
A recent study applied the VAE learning objective (ELBO), which generalises the softmax trained using the cross-entropy loss in supervised learning tasks \cite{Dhuliawala2023}. The variational classification latent variable model interprets the input to the final activation in neural network models as a sample of a Gaussian latent space to encourage more robust learning. The study found that the model maintained classification accuracy while improving model calibration, adversarial robustness and effectiveness in low-data settings. Further recent work has shown that VAEs are useful for other tasks, such as data augmentation where VAEs can generate realistic synthetic images from limited medical datasets \cite{rais2024exploringvariationalautoencodersmedical}. 

\section{Methodology}
\subsection{MRI Preprocessing and Augmentation}
Preprocessing was conducted using the TorchIO library \cite{torchio} to normalise the MRI scans and crop to ($32$ × $32$) voxels for each lymph node patch. The pipeline included histogram intensity scaling, Z normalisation, and resampling intensity values to [0,1] with a consistent scan orientation and a standard voxel size of ($0.573mm$ × $0.573mm$ × $3.3mm$). Each patch corresponds to a region of interest with an area of $336.2 mm^2$. The patches containing the largest LN segmentations were selected for each patient with a limit of 15, and empty spaces were filled with zero padding. To provide 3D information to the model, multiple 2D cross-sections of LNs were included if the segmentation maps of secondary patches covered at least half of the area. 

Data augmentation was used to increase dataset variability and to help mitigate the imbalanced and limited data. Spatial augmentations were horizontal and vertical flips, and translations to shift the patch centre and capture different context from the surroundings. Intensity-based augmentations applied included random Gaussian noise, gamma correction, and bias field distortion to simulate magnetic field inhomogeneities.

\subsection{VAE Model Description and Training}
\begin{figure*}[h!]
    \centering
    \includegraphics[width=\linewidth]{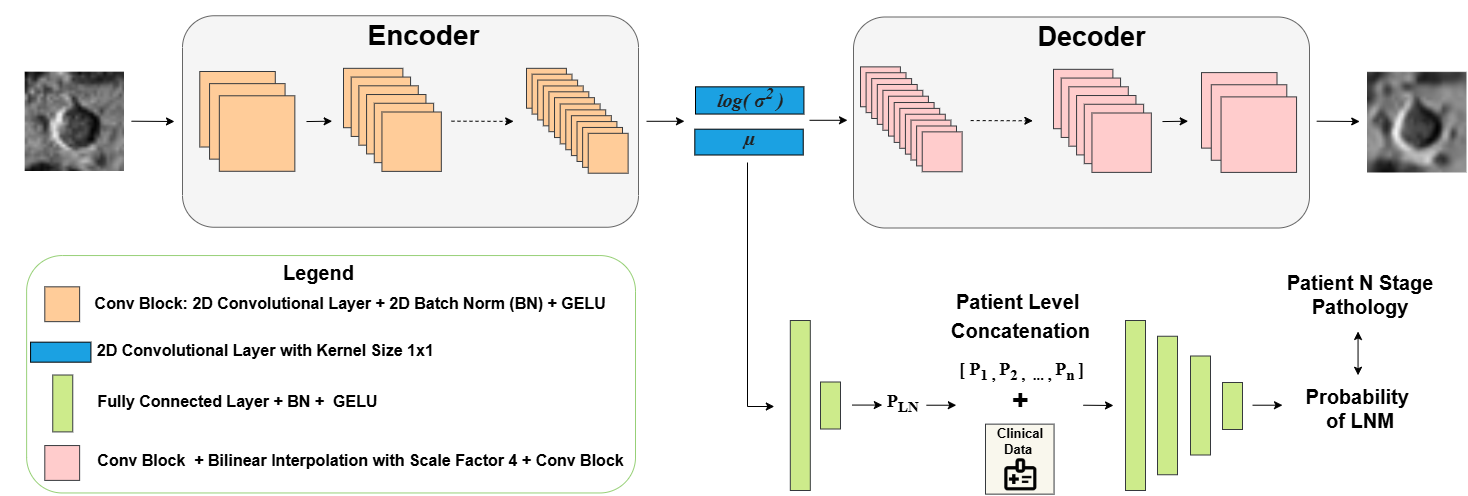}
    \caption{VAE-MLP model architecture}
    \label{fig:MIUA_model architecture}
\end{figure*}

\textbf{VAE Model Architecture. }
The proposed VAE encoder-decoder architecture is visualised in Fig.~\ref{fig:MIUA_model architecture}. 

The encoder has 6 blocks of 2D convolutional layers with 2D batch normalisation \cite{batch-norm}, and Gaussian Error Linear Unit (GELU) activation \cite{GELU}. The output of the encoder is used in two separate convolutional layers for the latent vectors, mean ($\mu$) and log variance $\left( \text{log}(\sigma^2) \right)$, forming a latent space of LN feature representations. The decoder takes a sample from the $n$-dimensional Gaussian distribution parameterised by the latent vectors, 
$z_i = \mu_i + \sigma_i \cdot \epsilon_i$, with noise sampled from a standard Gaussian, $\epsilon_{i} \sim \mathcal{N}(0,1)$.
The decoder is a symmetric architecture, with a bilinear interpolation in between the convolutional blocks to upsample the feature maps by a scale factor of 2 each time.
The final reconstructed images are evaluated against the original MRI patches in the loss function.
\newline \newline \textbf{Custom Loss Function.  }
The VAE loss function in Eq.~\ref{loss function} is a weighted combination of the L1 Loss (mean absolute error), the Kullback-Leibler Divergence (KLD) \cite{KL_divergence} and the Structural Similarity Index Measure (SSIM) \cite{ssim} for each image $i$ as follows,
\vspace{-1mm}
\begin{equation} \label{loss function}
    \mathcal {L}_{\text{VAE}} = \displaystyle\sum_{i=1}^n  \alpha \cdot \lambda \cdot \text{L1 Loss}_i + (1-\alpha) \cdot \gamma \cdot \text{SSIM}_i + a \cdot \beta \cdot \text{KLD}_i.
\end{equation}
The L1 Loss and SSIM measure image reconstruction quality and the KLD is the standard measure of latent space smoothness \cite{vae_original}. The metrics were averaged across the batch, and loss function hyperparameters $\alpha \in [0,1]$, $\lambda$, $\gamma$ and $\beta$ were included to help find a good balance of these adversarial metrics. 
An annealing function, `$a$', was included to exponentially increase the KLD throughout the training. This helps the VAE prioritise high-quality reconstructions in the early stages, before progressively enforcing a structured latent space. The KLD measures the divergence between the learned latent distribution and a standard Gaussian distribution. The loss function aims to balance high reconstruction quality with a latent space that generalises and disentangles the common features present in the data.
\newline \newline \textbf{VAE Model Training. }
The size of the VAE model was scaled by the hyperparameter `$\text{base}$' to control the number of feature maps, ranging from 16 to 28. In the encoder, the number of maps increased from $\text{base}$ to $16  \cdot  \text{base}$, with kernel sizes between 3 and 8. The latent size scalar was between 16 and 28, and the latent vector size was calculated as latent size $\cdot$ base, yielding a latent space dimensionality of 256 to 784. Training hyperparameters included the learning rate, weight decay, batch size, and the number of gradient accumulation steps (1 to 3). 
Models were trained for 200 epochs with early stopping checkpoints to avoid unnecessary computation.
To find the optimal model, we conducted a Bayesian search of the hyperparameter space with 200 runs. Results were stored and compared using the Python library Weights and Bias (WandB) \cite{wandb}.

The final VAE MRI reconstructions were evaluated qualitatively and quantitatively with the average SSIM, Mean Squared Error (MSE), Mean Absolute Error (MAE), Peak Signal-to-Noise Ratio (PSNR), and Learned Perceptual Image Patch Similarity (LPIPS) \cite{LPIPS_citation}. 

\subsection{Lymph Node Metastasis Classification}
\textbf{MLP Model Architecture. }
The classification MIL approach is visualised in Fig.~\ref{fig:MIUA_model architecture}. The MLP layers included 1D Batch Normalisation, dropout and GELU activation. The final layer of each MLP was a fully connected layer followed by a sigmoid activation. The model used VAE or CNN-derived features and selected clinical features in a simple MLP to make individual LN predictions. These predictions were combined with patient clinical data and input to a 4-layer MLP classification model for the final diagnosis. This model architecture was motivated by Xia et al. \cite{Xia2024_LNM} who successfully applied this setup for the same task and our experiments found that deeper networks did not improve the performance. 
\newline \newline \textbf{Clinical Features. }
Patient clinical data included age, sex, and the primary tumour stage (T1-T4). Size features, the long and short-axis diameter and their ratio, were measured using the segmentation mask and the standard voxel size. Border irregularity features included convexity, which describes how smooth the border is using the perimeter of the convex hull of the LN segmentation divided by the actual perimeter; and $\text{compactness} = \frac{\text{perimeter}^2}{4 \cdot \pi \cdot \text{area}}$ which measures how close the LN is to a perfect circle. These clinical features were normalised to the range $[0,1]$ using the min and max across the dataset. The size and border features were concatenated with the VAE or CNN feature representations before the first MLP. After making predictions for individual LNs, the patient clinical data and features for the largest short-axis LN were combined and input to the patient level classification model. 
\newline \newline \textbf{MLP Model Training. }
The final prediction is a weighted sum using $\eta \in [0.5,0.75]$ to scale the patient MLP prediction and $(1-\eta)$ to scale the maximum individual LN probability. The max is less impacted by noise and works best in cases where there is an obvious large malignant LN, whereas the patient MLP uses the clinical context and information from all detected LNs.

To help balance the dataset, between 5 and 30 synthetic patients were created by applying augmentations to the patches and oversampling positive patients in the training set. Other MLP hyperparameters included the learning rate, weight decay, batch size, dropout, and layer sizes in the MLP. These hyperparameters were optimised using a Bayesian search on WandB, aiming to improve the maximum test AUC. 
The optimal hyperparameter ranges were discovered through three Bayesian searches with 200 model runs, refining the space each time based on the highly selected parameters and the best results.
The final search found that 93/200 models achieved over 0.8 test AUC.

To provide a robust evaluation of the proposed VAE-MLP, we report a five-fold nested cross-validation (CV) score. Due to the limited data, a nested CV was used to optimise the model hyperparameters independently for each fold. This evaluation reflects the overall approach rather than a fixed set of hyperparameters. For each fold, the VAE with the highest test SSIM was selected from 5 candidates, followed by the MLP with the highest test AUC out of 10 candidates. These candidates were identified from the large hyperparameter searches. Final metrics were averaged over the five folds and given with the standard deviation to indicate the consistency of the performance.
\newline \newline \textbf{Ablation Study. } 
The key model components were tested using a strategy of switching one off at a time. Comparison models were trained on a narrow hyperparameter search with optimised ranges for 50 runs, selecting the highest test AUC. Firstly, the VAE backbone was compared with a CNN feature extractor. The architecture was a DenseNet, pre-trained on the ImageNet dataset \cite{imagenet} and using the implementation in MONAI \cite{monai}. Additionally, the model was tested without clinical data, and the final prediction using only the max LN prediction or only the patient MLP. 

\subsection{Post-Hoc Explainability}
Grad-CAM heatmaps were generated to show where the VAE model focuses when encoding the MRI patches. Gradients with respect to each pixel are averaged over 80 feature maps (8 × 8) from the last spatial layer,  and thresholded to display the top 25\%.

K-Means \cite{kmeans} clustering statistics are provided to justify that the latent space groups LNs based on information relevant to the diagnosis including size, shape and border irregularity. 
The number of clusters ($k$) was optimised to minimise the variability in lymph node size within the clusters. Additionally, clusters with less than 3 samples were excluded to provide more meaningful comparisons across clusters. Statistics calculated include the standard deviation and range of size and border irregularity metrics for LNs in the same cluster. They aim to describe how the MRI patches are clustered according to LN size and shape.

Lastly, we demonstrate that the latent space can simulate LN growth by applying the direction vector calculated between the average latent vectors of small LNs and large LNs.

\subsection{Dataset Description}
The study cohort included 168 RC patients who did not receive neo-adjuvant chemotherapy or radiotherapy treatment before a major resection surgery at Leeds Teaching Hospitals NHS Trust between 2010-2022. Ethical approval was given by the medical research ethics board of the author's institution, and  Health Research Authority (HRA) approval was obtained from the UK NHS ethics board (319850).
The pathology includes TNM staging for each patient where N0 represents no spread to the lymph nodes, N1 for 1-3 positive LNs, and N2 for 4 or more \cite{doi:10.1148/rg.2019180114}. The dataset was divided into training and test sets with a 65/35 split. This test set was made larger than usual due to the imbalanced classes.
The clinical and pathological summary of the study cohort is given in Table~\ref{table:dataset_characteristics_chapter3}.
\begin{table}[h!]
    \centering
    \caption{Dataset Characteristics for Train and Test Cohorts}
    \begin{tabular}{@{}l@{ }c@{ }c@{}}
        \toprule
        \textbf{Characteristic} & \textbf{  Train (n=109) } & \textbf{ Test (n=59)} \\ 
        \midrule
        Age (Mean ± $\sigma$)  & 69.7 ± 10.4 & 68.3 ± 13.0 \\ 
        Sex        & &  \\ 
        \hspace{5mm} Male                   & 80 (73.4) & 34 (57.6) \\
        \hspace{5mm} Female                 & 29 (26.1) & 25 (42.4) \\
        Number of Comorbidities & 2.4 ± 1.2 & 2.2 ± 1.2  \\ 
        Tumour Pathology  & & \\  
        \hspace{5mm} T1 & 19 (17.4) & 11 (18.6) \\
        \hspace{5mm} T2 & 52 (47.7) & 27 (45.8) \\
        \hspace{5mm} T3 & 32 (29.4) & 15 (25.4) \\
        \hspace{5mm} T4 & 6 (5.5) & 6 (10.2) \\
        Node Pathology & & \\     
        \hspace{5mm} N0 & 84 (77.1) & 46 (78.0) \\
        \hspace{5mm} N1 & 18 (16.5) & 8 (13.6) \\
        \hspace{5mm} N2 & 7 (6.4) & 5 (8.5) \\
        Number of Nodes on MRI &  &   \\ 
        (N (Mean ± $\sigma$)) & 625 (5.9 ± 3.8)\text{   } & \text{   }387 (5.9 ± 3.8)  \\ 
        Short-axis Diameter ($mm$)\text{  } & 5.6 ± 1.9 & 5.5 ± 1.9 \\
        Long-axis Diameter ($mm$) & 7.7 ± 2.6 & 7.7 ± 2.6 \\
        Short / Long-axis Ratio & 0.7 ± 0.1 & 0.7 ± 0.1 \\
        \bottomrule
    \end{tabular}
    \label{table:dataset_characteristics_chapter3}
\end{table}

The last scan before the surgery was selected for each patient and scans more than 16 weeks before the surgery were excluded as the TNM stage may change in the time between the MRI scan and the pathology. Of the 195 patients identified, some cases were excluded as there were no LNs detected in the axial scan (n=18), and others did not have an axial scan available (n=9).
\newline \newline \textbf{Lymph Node Annotations. } 
MRI segmentations of the LNs were completed by a science postgraduate student using the open-source software 3D Slicer version 5.4 \cite{3D_slicer,slicer_website}. 
To ensure annotation quality, a consultant GI radiologist provided training and reviewed a subset of 34 (20.2\%) MRI scans, focusing on challenging cases identified by the annotator and those frequently misclassified by the model.
Following the review, the total number of LNs in the sample changed from 303 to 271 with 46 removed and 14 added, this corresponds to a 94.8\% sensitivity in detecting LNs. Additionally, 17 scans (50\%) remained unchanged. 

The annotation process included detecting and segmenting LNs on the axial scan, followed by cross-referencing on the coronal and sagittal scans to update and finalise the segmentations. This task presented several challenges as LNs can be small and easily overlooked. Additionally, it can be difficult to distinguish LNs from other structures, most commonly blood vessels. However, blood vessels typically follow a trajectory through multiple slices, whereas LNs generally appear across 1-4 slices with a thickness of $3.3mm$. Other less common structures such as tumour deposits (TD), extramural vascular invasion (EMVI) and fibrotic tissue may also resemble LNs.

\section{Results}
\subsection{VAE MRI Lymph Node Patch Reconstruction Results}
\begin{figure}[ht]
    \centering
    \includegraphics[width=\linewidth]{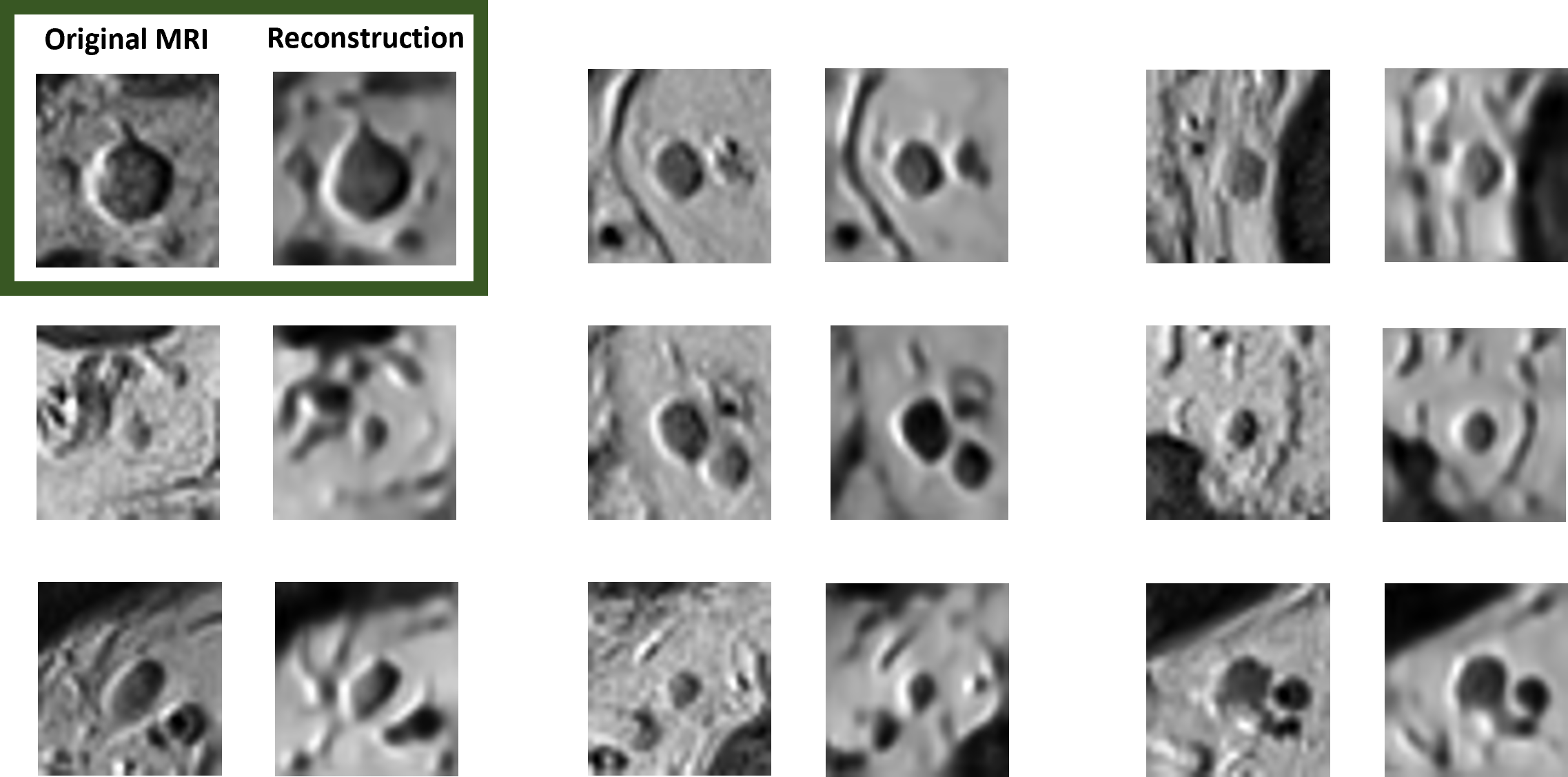}
    \caption{MRI lymph node patches and VAE reconstructions}
    \label{fig:vae_reconstructions}
\end{figure}

Firstly, a sample of images and VAE reconstructions are qualitatively reviewed in Fig.~\ref{fig:vae_reconstructions}. Observe that the key structures are captured very well although some of the smaller details are a little blurry.
Clinical collaborators in oncology
confirmed that the reconstructions captured important clinical features used in diagnosis, including size, shape and heterogeneous texture. The final VAE model image reconstruction metrics are given in Table~\ref{tab:reconstruction_metrics}.
\begin{table}[ht]
    \centering
    \caption{Image Reconstruction Metrics}
    \setlength{\tabcolsep}{6pt}
    \begin{tabular}{@{}lccccc@{}}
        \toprule
        \textbf{Dataset} & \textbf{SSIM} & \textbf{PSNR} & \textbf{LPIPS} & \textbf{MSE} & \textbf{MAE} \\ 
        \midrule
        Train & 0.796 & 26.971  & 0.0437 & 0.00203 & 0.0319 \\ 
        Test  & 0.789 & 26.596  & 0.0363 & 0.00235 & 0.0347 \\ 
        \bottomrule
    \end{tabular}
    \label{tab:reconstruction_metrics}
\end{table}

Optimal hyperparameters given here are averaged over the top 5 models based on SSIM, and rounded to the closest choice. Firstly, from the loss function (Eq.~\ref{loss function}): $\alpha = 0.5$, $\lambda = 4000$, $\gamma = 3 \cdot \text{batch size}$, and annealing = True. Secondly, training hyperparameters: base = 20, latent size = 20, learning rate $= 6.73 \cdot 10^{-4}$, weight decay = 0.035, batch size = 1024, and accumulation steps = 2.

\subsection{Classification Performance}
Model performance metrics are provided in Table~\ref{tab:results} for the best VAE-MLP model and the five-fold cross-validation score, the ablation study for the proposed model including a direct comparison to a CNN, and these results are compared with the metrics of four previous studies using external datasets. Model training graphs and the ROC curves are available in the supplementary materials and on GitHub. 

The most clinically important metric is sensitivity, as it represents how many positive cases were identified. Patients with LNM are usually recommended to have neo-adjuvant therapy to downstage the LNs before surgery. In this context, a lower sensitivity would correspond to more LNM patients advancing straight to surgery, potentially increasing the risk of recurrence and distant metastasis.
\begin{table*}[ht]
    \centering
    \caption{Classification performance metrics comparison. Showing the best VAE-MLP model and the cross-validation score (CV) in the first section, the ablation study results in the second, and comparisons with prior studies on external datasets in the third. Balanced Accuracy (BA), the average of sensitivity and specificity, is included to give an objective comparison metric across the different calibrations.}
    \resizebox{\textwidth}{!}{
    \begin{tabular}{@{}lcccccc@{}}
        \toprule
        \textbf{Model} & \textbf{AUC} & \textbf{Sensitivity} & \textbf{Specificity} 
        & \textbf{Accuracy} & \textbf{F1} & \textbf{BA}\\ 
        \midrule
         $\text{VAE-MLP}$  & \bf{0.888} & \bf{0.923} & \bf{0.848} & \bf{0.864} & \bf{0.733} & \bf{0.885} \\  
        $\text{VAE-MLP}_{\text{CV}}$ & $0.858 \pm 0.051$\text{ } & $0.789 \pm 0.064 $\text{ } & \text{ } $0.846 \pm 0.054 $\text{ } & $0.833 \pm 0.049 $\text{ } & 0.694 & 0.818 \\ 
        \midrule 
        $\text{CNN-MLP}$ & \bf{0.893} & \bf{0.846} & \bf{0.891} & \bf{0.881} & & \bf{0.869} \\
        $\text{VAE-MLP}_{\text{DL}}$ & 0.770 & 0.692 & 0.848 & 0.814 & & 0.770 \\ 
        $\text{VAE-MLP}_{\text{P}}$ & 0.847 & \bf{0.846} & 0.848 & 0.848 & & 0.847 \\ 
        $\text{VAE-MLP}_{\text{LN}}$ & 0.830 & 0.769 & \bf{0.891} & 0.864 & & 0.830 \\ 
        $\text{VAE-MLP}_{\text{Max}}$ & 0.771 & \bf{0.846} & 0.696 & 0.729 & & 0.771 \\ 
        $\text{VAE-MLP}_{\text{MLP}}$ & \bf{0.869} & \bf{0.846} & \bf{0.891} & \bf{0.881} & & \bf{0.869} \\
                \midrule
        Xia et al.~\cite{Xia2024_LNM}& $0.810 \pm 0.045$ & $0.702 \pm 0.072$ & 0.800  & & & 0.751 \\
        Wan et al.~\cite{Wan2023}& $0.790 \pm 0.100$ & \bf{1.000} & $0.660 \pm 0.110$ &	$0.730 \pm 0.090$ & & 0.830\\
        Liu et al.~\cite{Liu2023} & $\bf{0.942 \pm 0.055}$ & 0.955 & \bf{0.857} & \bf{0.895} & & \bf{0.906} \\ 
        Xie et al.~\cite{Xie2023} & $0.768 \pm 0.044$ & $0.763 \pm 0.056$ & $0.734 \pm 0.062$ & $0.747 \pm 0.012$ & 0.727 & 0.749 \\ 

        \bottomrule
    \end{tabular}}
    \label{tab:results}
\end{table*} 

The best model, $\text{VAE-MLP}$ uses a test set as large as feasible, with 59 patients representing 35\% of the 168 in total. The cross-validation (CV) average metrics are given with standard deviations, under $\text{VAE-MLP}_{\text{CV}}$. Note that the 20\% test set size had 7 or 8 positive cases, so each true positive impacted the sensitivity by at least 0.125. This setup constrains the representativeness and generalisability of individual fold models. The maximum fold performance had a sensitivity of 0.875 and specificity of 0.885, suggesting that some data splits contained easier cases. The cross-validation performance provides a robust measure of how the model performs on average. The performance metrics are still strong, indicating the approach is robust.
\newline \newline \textbf{Comparisons to Prior Work. }
Table~\ref{tab:results} displays four studies representing the current state-of-the-art. Firstly, to provide context to the results, \cite{Xia2024_LNM} is the most direct comparison as they used MRI and individual LN annotations, whereas \cite{Wan2023} and \cite{Liu2023} both used primary tumour annotations including only the closest LNs.  Since the LNs are small in comparison to the tumour this approach is indirect, and it is unclear if the model focuses on the LNs. In our dataset, you could get 0.605 sensitivity and 0.723 specificity just from assuming T1-2 is N0 and T3-4 is N1. Additionally, \cite{Liu2023} and \cite{Xie2023} used a CT dataset instead of MRI and are less direct comparisons.

Firstly, Xia et al. \cite{Xia2024_LNM} used a MIL framework with MRI annotations of individual LNs and a large CNN-MLP that incorporated LN size characteristics. They showed that senior radiologists could improve their AUC from 0.79 to 0.88 with model assistance. The model was deployed on an internal test cohort and two external validation cohorts, providing a good indication that the approach is generalisable. The best result from the internal test cohort is displayed in Table~\ref{tab:results}.

Wan et al. \cite{Wan2023} used a large pre-trained 3D ResNet on primary tumour MRI patches and reported results on a test set of 86 patients with early stage T1-2 rectal cancer.
They compared their results with three radiologists who had an average of 0.54 AUC. 

Liu et al. \cite{Liu2023} had the highest performance using a stacking nomogram which combined ResNet models and SVMs using radiomics features from the primary tumour on CT. However, the test set was relatively small (n=57), and the study inclusion criteria required more than 12 LNs examined pathologically, which likely translates to more visible LNs and cases that are less challenging to stage. Therefore while \cite{Liu2023} reports the best metrics, the study may lack generalisability. 

Lastly, Xie et al. \cite{Xie2023} used an attention-based MIL approach with a CNN and a logistic regression classifier. The performance is the lowest of the comparisons, however, it has the most robust validation with a thorough ablation study and is the only pre-operative colorectal cancer LNM study using deep learning to employ cross-validation.
\newline \newline \textbf{Optimal Hyperparameters. }
Classification model hyperparameters provided here are averaged over the top 10 performing models according to the maximum test AUC score and rounded to the closest choice: $\eta = 0.8$, number of synthetic patients = 25, oversample ratio for positive cases = 1.5, and the number of slices per patient = 20. Training and model hyperparameters were: threshold = 0.436, batch size = 128, learning rate $= 6.59 \cdot 10^{-3}$, accumulation steps = 4, weight decay = 0.16, patch hidden dimension = 2048, patient hidden dim = 96, patch dropout = 0.4, and patient dropout = 0.3.

\subsection{Ablation Study}
Table~\ref{tab:results} displays the ablation study results from testing key model components using a narrow hyperparameter search of 50 runs. Firstly, we provide a baseline comparison of the VAE with the standard approach of using a CNN feature extractor. The CNN-MLP model replaces the VAE encoder with a pre-trained DenseNet and the same classification head as the VAE-MLP model. The feature size was 1024, much larger than the final VAE latent size of 400. The results show that the CNN had lower performance metrics than the VAE.

The ablation of the VAE-MLP checks that key components are improving the overall performance. First, the model with just deep learning features ($\text{VAE-MLP}_{\text{DL}}$) had the worst performance, justifying the use of clinical data. Then two models using just the T stage and age/sex ($\text{VAE-MLP}_{\text{P}}$), and just the LN size and border metrics ($\text{VAE-MLP}_{\text{LN}}$). The patient level data appears to be more important in terms of AUC, and the use of both sets of clinical features had the best performance. Lastly, we tested the MIL methods of aggregation, one model used only the max LN prediction ($\text{VAE-MLP}_{\text{Max}}$) and another with only the patient MLP ($\text{VAE-MLP}_{\text{MLP}}$). Overall, the patient MLP had the best performance, however a combination including both the patient MLP and the max value was used in the proposed VAE-MLP as the max is less impacted by noise and imbalanced data. 
 
\subsection{Error Analysis and Uncertainty}
Predictions were accumulated from 500 model states saved across the hyperparameter runs when the performance for one epoch was above 0.8 test AUC. Note that this included multiple predictions from the same run, whenever the threshold was met. An error graph displaying these results is available in the supplementary materials and on GitHub. It shows that 11 patients in the test set were misclassified in over half of the saved results, clearly indicating they are the most difficult cases. Upon inspection of the specific examples, the senior radiologist confirmed they were difficult to stage, and identified small or medium suspicious LNs with an irregular border and/or heterogeneous texture.
We found that the frequent false positive (FP) cases all had a large LN with a short-axis diameter between 7-10$mm$. Frequent false negative (FN) cases had 5.6 LNs on average, much less than 9.2 on average for all positive cases. The largest short-axis diameter LN per patient was 39\% higher on average for the frequent FPs than for the frequent FN cases. 

To provide a measure of uncertainty, average prediction confidence was accumulated across the 500 model states. In the test set, there was a minimum average prediction of 0.15 and a maximum of 0.78. If we define a prediction in the range 0.33 to 0.66 as uncertain, then 19 patients (32.2\% of the test set) had an uncertain diagnosis. Of the common FNs, the average confidence was 0.23 and the FPs had an average of 0.50 further indicating the uncertainty. In clinical applications, we would recommend an expert review of these cases.

\subsection{Interpretability and Clustering}
\begin{figure}[h!]
    \centering
    \includegraphics[width=0.46\linewidth]{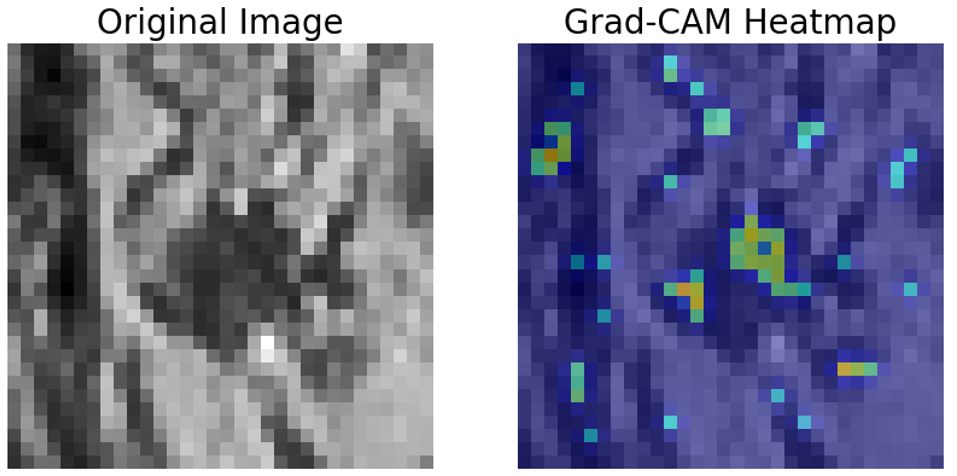}
    \hspace{0.02\linewidth}
    \includegraphics[width=0.46\linewidth]{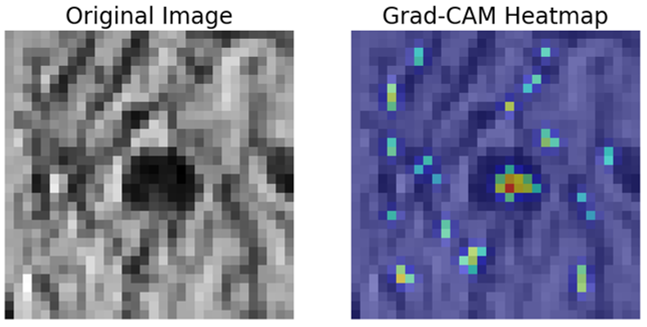}
    \caption{Grad-CAM heatmaps showing gradients from the last spatial layer of the encoder before flattening.}
    \label{fig:gradcam}
\end{figure}

Grad-CAM examples in Fig.~\ref{fig:gradcam} show that the VAE encoder is focusing on the LNs and important visual features when encoding the MRI patches.

Clustering statistics are shown in Table~\ref{tab:clustering_stats}, demonstrating that the feature representations of individual LNs are separated by clinically important criteria including the LN size features and the average of the border irregularity (BI) metrics. The optimal number of clusters was around 35, and the global standard deviations ($\sigma$) and ranges for the features across all LNs are provided as a baseline for comparison.  
These results aim to describe the average variability within clusters. The standard deviation and ranges indicate that variability within the clusters is lower than the global variability, suggesting that patches are grouped based on similar LN size and shape. 
\begin{table}[ht]
    \centering
    \caption{Clustering Statistics for short and long axis diameter given in ($mm$), the ratio short to long, and the average of compactness and convexity for border irregularity metrics in the range $[0,1]$. The statistics given aim to describe the inter-cluster variations using the standard deviation ($\sigma_{\text{IC}}$) and range (Range${}_{\text{IC}}$).}
    \begin{tabular}{@{}lcccc@{}}
        \toprule
        \textbf{Statistic} & \textbf{Short-axis  } & \textbf{Long-axis   } & \textbf{Ratio } & \textbf{ BI} \\
        \midrule    
        Global $\sigma$ & 2.13 & 2.79 & 0.12 & 0.13 \\ 
        Mean $\sigma_{\text{IC}}$ & 1.815 & 2.522 & 0.119 & 0.118 \\
        Min $\sigma_{\text{IC}}$ & 0.696 & 1.282 & 0.068 & 0.058 \\
        Max $\sigma_{\text{IC}}$ & 3.222 & 4.206 & 0.334 & 0.190 \\
        \midrule
        Global Range & 13.167 & 16.145 & 0.811 & 1 \\
        Mean Range${}_{\text{IC}}$ & 7.290 & 9.616 & 0.431 & 0.346 \\
        Min Range${}_{\text{IC}}$ & 1.717 & 3.022 & 0.169 & 0.093 \\
        Max Range${}_{\text{IC}}$ & 12.721 & 15.929 & 0.811 & 0.678 \\
        \bottomrule
    \end{tabular}
    \label{tab:clustering_stats}
\end{table}

\begin{figure}[h!]
    \vspace{-1mm}
    \centering
    \begin{subfigure}{\linewidth}
        \centering
        \includegraphics[width=\linewidth]{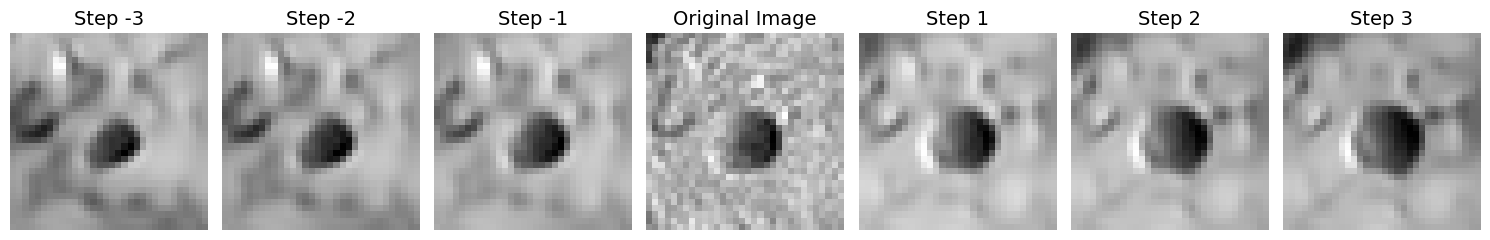}
    \end{subfigure}
    \begin{subfigure}{\linewidth}
        \centering
        \includegraphics[width=\linewidth]{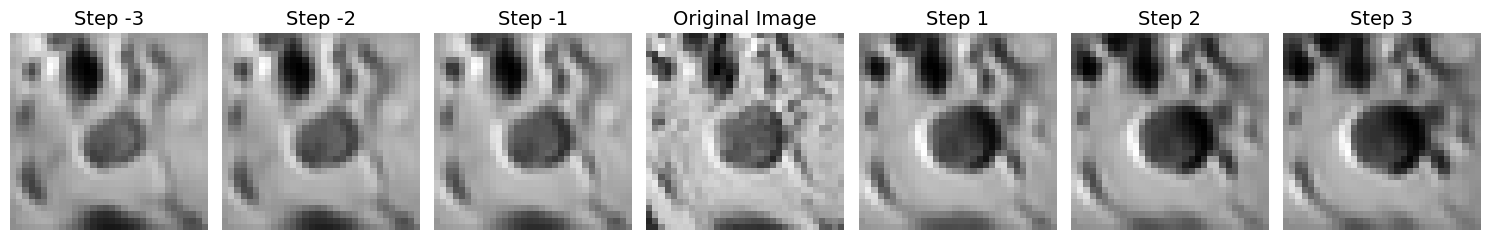}
    \end{subfigure}
    \caption{Lymph node growth direction, with the original image in the centre and reconstructed images after applying multiples of the direction vector on either side.}
    \label{fig:traversals}
\end{figure}

Lastly, Fig.~\ref{fig:traversals} shows two examples that apply the average direction gradient between small and large LNs in the latent space. There is a smooth evolution of LN growth whilst the background is fixed, demonstrating that the latent space has generalised and disentangled lymph node growth as an independent factor.

\section{Discussion}
Some key methodological considerations are covered here to inform future work. Firstly, extending the analysis to 3D would provide more information about the geometric shape of the lymph nodes and more context with the surroundings. Secondly, multi-class LN classification may be beneficial in future work as it can indicate the need for different levels of neo-adjuvant treatment, although an accurate binary diagnosis is the most important result. Third, we note that it is not feasible to match LNs detected on MRI with the pathologic status of individual LNs in retrospective studies due to the difficulty of matching them post-surgery. Only two previous studies have done this, with an obvious selection bias as they only matched the pathology of the largest detected LNs \cite{Li2021_LNM,Ozaki2023}. Fourth, all LNs identified in the axial scans were included, although the mesorectal LNs have a higher probability of metastasis. This presents a partial limitation as including more normal LNs introduces more noise into the classification model. Future work might improve on this by identifying an approximate location of the primary tumour and measuring the distance, then a classifier could weight the contribution of LNs based on proximity to the tumour. 
Lastly, the annotations were completed by a non-specialist, potentially leading to some LNs being omitted and other non-LN structures being included. However, 20\% of the annotations were validated by a consultant GI radiologist who confirmed the annotations were of sufficient quality. Also, the N stage pathology is used to evaluate the predictions and so a high classification performance indicates that the annotations are accurate. The use of lower quality annotations in this task may be beneficial for future work to reduce the study development costs of using large datasets. To enhance model explainability, future work may look at using the VAE latent space to provide prototype-based or counterfactual explanations.

\section{Conclusion}
In conclusion, this paper provides a pilot study for the radiological staging of rectal cancer lymph node metastasis on MRI using VAEs. We have observed that the latent space is well structured for the classification task and encodes clinically meaningful characteristics. The best result of 0.89 AUC, 92.3\% sensitivity and 84.8\% specificity has exceeded the performance of the most robust studies in the literature, and the average radiologist performance of 73\% sensitivity and 74\% specificity \cite{Zhuang2021}. The cross-validation, ablation study, error analysis and interpretability methods further validate the model approach and performance. This study covers 12 years of clinical practice where 38 patients had lymph node metastasis but were not referred for neo-adjuvant treatment which may have benefited them, our model could have identified up to 92.3\% of these cases.

\begin{credits}
\subsubsection{\ackname} Benjamin Keel is supported by the EPSRC Centre for Doctoral Training in Artificial Intelligence for Medical Diagnosis and Care (EP/S024336/1).
\end{credits}

\bibliographystyle{splncs04}
\bibliography{bibliography}

\begin{thebibliography}{10}
\providecommand{\url}[1]{\texttt{#1}}
\providecommand{\urlprefix}{URL }
\providecommand{\doi}[1]{https://doi.org/#1}

\bibitem{nhs_cancer_stats}
Nhs digital cancer survival in england, cancers diagnosed 2016 to 2020 (2023), \url{https://digital.nhs.uk/data-and-information/publications/statistical/cancer-survival-in-england/cancers-diagnosed-2016-to-2020-followed-up-to-2021/}

\bibitem{wandb}
Biewald, L.: Experiment tracking with weights and biases (2020), \url{https://www.wandb.com/}

\bibitem{Borgheresi2022}
Borgheresi, A., {De Muzio}, F., Agostini, A., Ottaviani, L., Bruno, A., et~al.: {Lymph Nodes Evaluation in Rectal Cancer: Where Do We Stand and Future Perspective}. Journal of Clinical Medicine  \textbf{11}(9),  1--26 (2022). \doi{10.3390/jcm11092599}

\bibitem{monai}
Cardoso, M.J., Li, W., Brown, R., Ma, N., Kerfoot, E., et~al.: Monai: An open-source framework for deep learning in healthcare (2022), \url{https://arxiv.org/abs/2211.02701}

\bibitem{imagenet}
Deng, J., Dong, W., Socher, R., Li, L.J., Li, K., Fei-Fei, L.: Imagenet: A large-scale hierarchical image database. In: 2009 IEEE Conference on Computer Vision and Pattern Recognition. pp. 248--255 (2009). \doi{10.1109/CVPR.2009.5206848}

\bibitem{Dhuliawala2023}
Dhuliawala, S., Sachan, M., Allen, C.: {Variational Classification} pp. 1--17 (2023), \url{http://arxiv.org/abs/2305.10406}

\bibitem{3D_slicer}
Fedorov, A., Beichel, R., Kalpathy-Cramer, J., Finet, J., Fillion-Robin, J.C., et~al.: 3d slicer as an image computing platform for the quantitative imaging network. Magnetic Resonance Imaging  \textbf{30}(9),  1323--1341 (November 2012), \url{https://www.ncbi.nlm.nih.gov/pmc/articles/PMC3466397/pdf/nihms383480.pdf}

\bibitem{GELU}
Hendrycks, D., Gimpel, K.: Gaussian error linear units (gelus)  (2016). \doi{10.48550/ARXIV.1606.08415}, \url{https://arxiv.org/abs/1606.08415}

\bibitem{doi:10.1148/rg.2019180114}
Horvat, N., Carlos Tavares~Rocha, C., Clemente~Oliveira, B., Petkovska, I., Gollub, M.J.: Mri of rectal cancer: Tumor staging, imaging techniques, and management. RadioGraphics  \textbf{39}(2),  367--387 (2019). \doi{10.1148/rg.2019180114}, \url{https://doi.org/10.1148/rg.2019180114}, pMID: 30768361

\bibitem{10.1093/bioadv/vbac100}
Hsu, T.C., Lin, C.: Learning from small medical data—robust semi-supervised cancer prognosis classifier with bayesian variational autoencoder. Bioinformatics Advances  \textbf{3}(1),  vbac100 (01 2023). \doi{10.1093/bioadv/vbac100}, \url{https://doi.org/10.1093/bioadv/vbac100}

\bibitem{batch-norm}
Ioffe, S., Szegedy, C.: Batch normalization: Accelerating deep network training by reducing internal covariate shift. CoRR  (2015), \url{http://arxiv.org/abs/1502.03167}

\bibitem{Kaur2023}
Kaur, G., Keshta, I., Shabaz, M., Batra, H.S., {Vijaya Sagar}, T., et~al.: {Rectal Cancer Prediction and Performance Based on Intelligent Variational Autoencoders Machine Using Deep Learning on CDAS Dataset}. Journal of Artificial Intelligence and Technology  \textbf{3}(4),  195--204 (2023). \doi{10.37965/jait.2023.0241}

\bibitem{VAE_lung_lesion}
Keel, B., Quyn, A., Jayne, D., Relton, S.D.: Variational autoencoders for feature exploration and malignancy prediction of lung lesions. In: British Machine Vision Conference (BMVC) (2023). \doi{10.48550/arXiv.2311.15719}, \url{https://doi.org/10.48550/arXiv.2311.15719}

\bibitem{keel2024state-of-the-art}
Keel, B., Quyn, A., Jayne, D., Relton, S.D.: State-of-the-art performance of deep learning methods for pre-operative radiologic staging of colorectal cancer lymph node metastasis: a scoping review. BMJ Open  \textbf{14}(12) (2024). \doi{10.1136/bmjopen-2024-086896}, \url{https://bmjopen.bmj.com/content/14/12/e086896}

\bibitem{vae_original}
Kingma, D.P., Welling, M.: Auto-encoding variational bayes. arXiv  (2013). \doi{10.48550/ARXIV.1312.6114}, \url{https://arxiv.org/abs/1312.6114}

\bibitem{KL_divergence}
Kullback, S., Leibler, R.A.: On information and sufficiency. The Annals of Mathematical Statistics  \textbf{22}(1),  79--86 (1951), \url{http://www.jstor.org/stable/2236703}

\bibitem{Li2021_LNM}
Li, J., Zhou, Y., Wang, P., Zhao, H., Wang, X., et~al.: {Deep transfer learning based on magnetic resonance imaging can improve the diagnosis of lymph node metastasis in patients with rectal cancer}. Quantitative Imaging in Medicine and Surgery  \textbf{11}(6),  2477--2485 (2021). \doi{10.21037/qims-20-525}

\bibitem{Liu2023}
Liu, J., Sun, L., Lu, X., Geng, Y., Zhang, Z.: Development and validation of a stacking nomogram for predicting regional lymph node metastasis status in rectal cancer via deep learning and hand-crafted radiomics. International Journal of Radiation Research  \textbf{21}(2) (2023). \doi{10.52547/ijrr.21.2.13}, \url{http://ijrr.com/article-1-4723-en.html}

\bibitem{kmeans}
MacQueen, J.: Some methods for classification and analysis of multivariate observations (1967)

\bibitem{Ozaki2023}
Ozaki, K., Kurose, Y., Kawai, K., Kobayashi, H., Itabashi, M., et~al.: {Development of a Diagnostic Artificial Intelligence Tool for Lateral Lymph Node Metastasis in Advanced Rectal Cancer}. Diseases of the Colon and Rectum  \textbf{66}(12),  E1246--E1253 (2023). \doi{10.1097/DCR.0000000000002719}

\bibitem{torchio}
P{\'e}rez-Garc{\'i}a, F., Sparks, R., Ourselin, S.: {TorchIO}: a {Python} library for efficient loading, preprocessing, augmentation and patch-based sampling of medical images in deep learning. Computer Methods and Programs in Biomedicine p. 106236 (2021). \doi{https://doi.org/10.1016/j.cmpb.2021.106236}, \url{https://www.sciencedirect.com/science/article/pii/S0169260721003102}

\bibitem{rais2024exploringvariationalautoencodersmedical}
Rais, K., Amroune, M., Benmachiche, A., Haouam, M.Y.: Exploring variational autoencoders for medical image generation: A comprehensive study (2024), \url{https://arxiv.org/abs/2411.07348}

\bibitem{Redekop_2024_CVPR}
Redekop, E., Pleasure, M., Wang, Z., Sarma, K.V., Kinnaird, A., et~al.: Codebook vq-vae approach for prostate cancer diagnosis using multiparametric mri. In: Proceedings of the IEEE/CVF Conference on Computer Vision and Pattern Recognition (CVPR) Workshops. pp. 2365--2372 (June 2024)

\bibitem{Selvaraju_2019}
Selvaraju, R.R., Cogswell, M., Das, A., Vedantam, R., Parikh, D., Batra, D.: Grad-{CAM}: Visual explanations from deep networks via gradient-based localization. International Journal of Computer Vision  \textbf{128}(2),  336--359 (oct 2019). \doi{10.1007/s11263-019-01228-7}

\bibitem{slicer_website}
Slicer, D.: 3d slicer: An open source software platform for medical image computing and research. \url{https://www.slicer.org/} (2024), accessed: 2024-10-24

\bibitem{Sreelekshmi2024}
Sreelekshmi, V., Pavithran, K., Nair, J.J.: Unleashing the power of hierarchical variational autoencoder for predicting breast cancer. IEEE Access  \textbf{12},  195658--195670 (2024). \doi{10.1109/ACCESS.2024.3518612}

\bibitem{valentini2011nomograms}
Valentini, V., van Stiphout, R.G., Lammering, G., Gambacorta, M.A., Barba, M.C., et~al.: Nomograms for predicting local recurrence, distant metastases, and overall survival for patients with locally advanced rectal cancer on the basis of european randomized clinical trials. Journal of Clinical Oncology  \textbf{29}(23),  3163--3172 (2011). \doi{10.1200/JCO.2010.33.1595}, \url{https://ascopubs.org/doi/abs/10.1200/JCO.2010.33.1595}

\bibitem{Wan2023}
Wan, L., Hu, J., Chen, S., Zhao, R., Peng, W., et~al.: {Prediction of lymph node metastasis in stage T1–2 rectal cancers with MRI-based deep learning}. European Radiology pp. 3638--3646 (2023). \doi{10.1007/s00330-023-09450-1}, \url{https://doi.org/10.1007/s00330-023-09450-1}

\bibitem{10577275}
Wang, J., Li, J., Wang, R., Zhou, X.: Vae-driven multimodal fusion for early cardiac disease detection. IEEE Access  \textbf{12},  90535--90551 (2024). \doi{10.1109/ACCESS.2024.3420444}

\bibitem{ssim}
Wang, Z., Bovik, A.C., Sheikh, H., Simoncelli, E.: Image quality assessment: from error visibility to structural similarity. IEEE Transactions on Image Processing  \textbf{13}(4),  600--612 (2004), \url{https://doi.org/10.1109/TIP.2003.819861}

\bibitem{Xia2024_LNM}
Xia, W., Li, D., He, W., Pickhardt, P.J., Jian, J., et~al.: {Multicenter Evaluation of a Weakly Supervised Deep Learning Model for Lymph Node Diagnosis in Rectal Cancer at MRI}. Radiology: Artificial Intelligence  \textbf{6}(2) (2024). \doi{10.1148/ryai.230152}

\bibitem{Xie2023}
Xie, M., Zhang, Y., Li, X., Mao, Y., Zou, X., Zhang, H.: {Predicting Lymph Node Metastasis of Colorectal Cancer in CT Scans Using Attention-Based Multiple Instance Learning}. 2023 IEEE International Conference on Bioinformatics and Biomedicine (BIBM) pp. 2695--2701 (2023). \doi{10.1109/BIBM58861.2023.10385936}

\bibitem{LPIPS_citation}
Zhang, R., Isola, P., Efros, A.A., Shechtman, E., Wang, O.: The unreasonable effectiveness of deep features as a perceptual metric. In: 2018 IEEE/CVF Conference on Computer Vision and Pattern Recognition. pp. 586--595 (2018). \doi{10.1109/CVPR.2018.00068}

\bibitem{Zhuang2021}
Zhuang, Z., Zhang, Y., Wei, M., Yang, X., Wang, Z.: {Magnetic Resonance Imaging Evaluation of the Accuracy of Various Lymph Node Staging Criteria in Rectal Cancer: A Systematic Review and Meta-Analysis}. Frontiers in Oncology  \textbf{11}(July) (2021). \doi{10.3389/fonc.2021.709070}

\end{thebibliography}

\end{document}